\documentclass[12pt,titlepage]{article}
\usepackage[letterpaper,top=1in,bottom=1in,left=1.25in,right=1.25in]{geometry}
\usepackage{graphicx,cite, url}
\usepackage{color}
\usepackage{hyperref}
\hypersetup{
    colorlinks=true,
    linkcolor=blue,
    filecolor=magenta,      
    urlcolor=blue,
}
\newcommand*{\img}[1]{%
    \raisebox{-.2\baselineskip}{%
        \includegraphics[
        height=\baselineskip,
        width=\baselineskip,
        keepaspectratio,
        ]{#1}%
    }%
}
\newcommand{\ft}{\mathcal{F}} 

\usepackage{amsmath,mathrsfs,cleveref}

\begin{document}

\title{Deep learning approach to Fourier ptychographic microscopy}

\author{Thanh Nguyen$^{1,*}$, Yujia Xue$^{2}$, Yunzhe Li$^{2}$, Lei Tian$^{2,*}$, George Nehmetallah$^{1}$
\\
\multicolumn{1}{p{\textwidth}}{\centering\emph{\normalsize 1. The Catholic University of America, 620 Michigan NE, Washington, DC 20064, USA\\
		2. Department of Electrical and Computer Engineering, Boston University, Boston, MA 02215, USA\\
		$^{*}$ 32nguyen@cua.edu\\
		$^{*}$ lei\_tian@alum.mit.edu
}}}

\maketitle
\begin{abstract}

Convolutional neural networks (CNNs) have gained tremendous success in solving complex inverse problems.
The aim of this work is to develop a novel CNN framework to reconstruct video sequence of dynamic live cells captured using a computational microscopy technique, Fourier ptychographic microscopy (FPM). 
The unique feature of the FPM is its capability to reconstruct images with both wide field-of-view (FOV) and high resolution, i.e. a large space-bandwidth-product (SBP), by taking a series of low resolution intensity images.
For live cell imaging, a single FPM frame contains thousands of cell samples with different morphological features.  
Our idea is to fully exploit the statistical information provided by this large spatial ensembles so as to make predictions in a sequential measurement, without using any additional temporal dataset.  
Specifically, we show that it is possible to reconstruct high-SBP  dynamic cell videos by a CNN trained only on the first FPM dataset captured at the beginning of a time-series experiment.  
Our CNN approach reconstructs a 12800$\times$10800 pixels phase image using only $\sim$25 seconds, a 50$\times$ speedup compared to the model-based FPM algorithm. 
In addition, the CNN further reduces the required number of images in each time frame by $\sim6\times$.  
Overall, this significantly improves the imaging throughput by reducing both the acquisition and computational times.  
The proposed CNN is based on the conditional generative adversarial network (cGAN) framework.  We further propose a mixed loss function that combines the standard image domain loss and a weighted Fourier domain loss, which leads to improved reconstruction of the high frequency information. Additionally, we also exploit transfer learning so that our pre-trained CNN can be further optimized to image other cell types. 
Our technique demonstrates a promising deep learning approach to continuously monitor large live-cell populations over an extended time and gather useful spatial and temporal information with sub-cellular resolution.
\end{abstract}

\maketitle

\section{Introduction}
In recent years, {\it data-driven} image reconstruction techniques based on machine learning, in particular {deep learning} (DL)~\cite{LeCun2015}, have gained tremendous success in solving complex inverse problems~\cite{lucas2018using}, and can often provide results surpassing those using state-of-the-art {\it model-based} techniques.  Traditionally, solving an inverse problem involves first explicitly formulating the imaging model and incorporating domain and prior knowledge (e.g. via the use of regularization techniques), and then finding an analytical solution (e.g. through an optimization procedure)~\cite{bertero1998introduction}.  Unlike model-based approaches, the `end-to-end' DL framework does not explicitly utilize any models or priors, and instead relies on large datasets to `learn' the underlying inverse problem.  The outcome of this DL approach consists of two important components.  First, the result from the training stage is a CNN that corresponds to {\it a} plausible underlying mapping function relating the measurement to the solution.  Second, the trained CNN can be used to make `predictions' when presenting it with new measurements that were unused in the training stage.  This second part comes with major practical benefits in computational cost and speed in typical image reconstruction problems, since the prediction process simply involves the feedforward computation of the CNN that typically takes no more than a few seconds on a normal grade GPU.  In contrast, most of modern model-based techniques rely on iterative algorithms~\cite{Afonso.etal2010,Beck.Teboulle2009,admm2011} that require much higher computational cost and longer running time; the same lengthy process needs to be repeated every time for each new measurement. 

Here, we distinguish two classes of imaging problems: those involve {\it independent} datasets from often static objects, and those dealing with {\it sequential} datasets that are temporally correlated, from dynamic objects.  In {\it independent}  problems, CNNs have been demonstrated to provide superior performance to solve many challenging imaging problems, such as image super-resolution~\cite{ledig2016photo,rivenson2017deep}, denoising~\cite{Burger.etal2012,zhang2017beyond},  segmentation~\cite{ronneberger2015u},  deconvolution~\cite{xu2014deep, zeiler2010deconvolutional}, compressive imaging~\cite{yao2017dr,kulkarni2016reconnet}, tomography~\cite{jin2017deep,nguyen2018com}, digital labeling~\cite{christiansen2018silico}, holography~\cite{rivenson2018phase,ren2018learning}, phase recovery~\cite{sinha2017lensless}, and imaging through diffusers~\cite{Li_2018,Li:2018aa}.  What's common in this class of problems is that {\it independently} prepared input-output pairs (i.e. measurement and solution), obtained by repeating the same imaging process, are presented to the CNN at the training stage to optimize the network's parameters.  
In  {\it sequential} problems,  the temporal correlation of a dynamic process contains additional information, and is often recorded in video datasets.  Various CNN frameworks have been proposed to learn the additional temporal information.  For example, spatial super-resolution has been demonstrated by training a CNN on both spatial and temporal dimensions of videos~\cite{kappeler2016video}.  Temporal super-resolution on recurring processes is achieved by learning the underlying temporal statistics~\cite{shahar2011space}. The motion information of dynamic objects is learned with an optical-flow based CNN~\cite{dosovitskiy2015flownet}.  Motion artifacts can be removed by jointly learning the blurring point-spread-function (PSF) and deconvolution operation~\cite{nah2017deep,chen2018reblur2deblur}.  In all these cases, CNNs are designed to process {\it a video sequence} in order to extract the temporal information.  The downside is that the CNN architectures inevitably become more complicated that require more computational resources, as compared to those used in the independent problems.  Fundamentally, the complication stems from that any single frame from the imaging techniques used does not contain sufficient temporal statistical information.

\begin{figure}[t]
\centering\includegraphics[width=\linewidth]{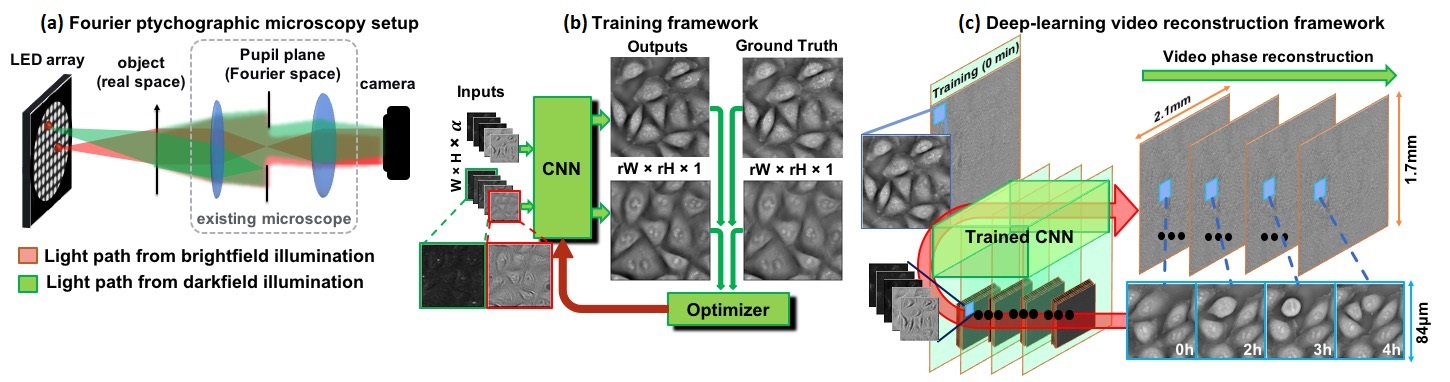}
\caption{The workflow of the proposed deep learning based Fourier ptychography video reconstruction. (A) The intensity data is captured by illuminating the sample from different angles with an LED array. (B) Training CNN to reconstruct high-resolution phase images. The input to the CNN are low-resolution intensity images; the output of the CNN is the ground truth phase image reconstructed using the traditional FPM algorithm in~\cite{Tian2015b}. The network is then trained by optimizing network's parameters that minimizes a loss function calculated based on the network's predicted output and the ground truth. (C) The network is fully trained using the first dataset at 0 min, then can be used to predict phase videos of dynamic cell samples frame by frame.}
\label{overview}
\end{figure}

In this work, we develop a CNN architecture to reconstruct video sequence of dynamic live cells captured with a computational microscopy technique based on Fourier ptychographic microscopy (FPM)~\cite{Zheng2013, Tian2015b}.  The unique feature of the FPM is its ability to quantitatively reconstruct phase information with both wide field-of-view (FOV) and high spatial resolution, i.e., a large space-bandwidth product (SBP).  This is not possible for traditional  techniques which must trade spatial or temporal resolution for FOV. 
For live-cell imaging applications, this allows one to simultaneously image a large cell population (e.g. more than 3400 in a single frame in ~\cite{Tian2015b}).  Cells of the same type undergo similar morphological changes during different cell states, which then repeat over each cell cycle.  If one records only a few cells at a time using conventional microscopy techniques~\cite{Stephens2003}, capturing the full dynamics would require a large sequence of measurements to cover the entire cell cycle (typically ranging from a few hours to days).  Our proposed technique is based on the observation that, in any live cell experiment without precise cell synchronization~\cite{ashihara197920}, at any instant of time, a large cell population would contain samples covering all cell states.  {\it In other words, it is possible to  gather sufficient temporal statistical information of a single cell by imaging a large spatial ensembles simultaneously}. Based on this idea, we propose a CNN that is trained using only a single frame from the FPM.  We then show that this trained CNN is able to reconstruct large-SBP phase videos with high fidelity using datasets taken in a time-series live cell experiments.  

Existing FPM techniques are limited by their long acquisition times, which are  limited by the FPM algorithms that require at least $65\%$ overlap in the Fourier coverage of the images captured from neighboring LEDs~\cite{Zheng2013}. Several illumination multiplexing techniques have been demonstrated to improve the acquisition speed~\cite{Tian2014,Tian2015b}. However, the amount of data reduction is still  limited by the Fourier overlap requirement.  Here, we show that, similar to prior work on CNN for FPM  on static objects~\cite{kappeler2017ptychnet}, our CNN can be sufficiently trained using much fewer images than that needed by the model-based FPM algorithms for dynamic live-cell samples. 

Distinct from computer vision applications, a particular challenge in applying DL to biomedical microscopy is the difficulty in gathering ground truth data needed for training the network. Various strategies have been proposed, including synthetic data from simulations built with physical imaging models~\cite{nehme2018deep,weigert2017content,boyd2018deeploco}, semi-synthetic data that uses experimental data to guide simulations~\cite{weigert2017content}, experimental data captured with a different modality~\cite{rivenson2017deep,rivenson2018phase}, and experimental data captured with the same modality~\cite{weigert2017content}.  Here, we propose to use the traditional FPM reconstructed phase images as the ground truth for training.  Since our technique requires only a single frame for training, this does not add much overhead in data acquisition or computation.  
When using experimental data as the ground truth, they inevitably are contaminated with noise. In FPM, the quality of the phase reconstruction is limited by spatially variant aberrations, system mis-alignment, and intensity-dependent noise~\cite{Yeh2015}.  Robust learning using noisy labelled data has been demonstrated for image classification and segmentation~\cite{xiao2015learning, lu2017learning}.  In essence, CNN captures the invariants while filtering out the random fluctuations~\cite{bengio2013representation, zeiler2014visualizing}.  Here, we show that our proposed CNN is also robust to phase noise in the `ground truth' data for solving the inverse problem of FPM.

We build a CNN based on the conditional generative adversarial network (cGAN) framework, consisting of two sub-networks, the generator and the discriminator. 
 The generator network uses the UNet architecture ~\cite{ronneberger2015u} with densely connected convolutional blocks (DenseNet)~\cite{huang2017densely} to output high-resolution phase image.  
The discriminator network distinguishes if the output is real or fake.  
We compare five variants of the network, which differ by the input measurements using different illumination patterns corresponding to different Fourier coverages. 
 Similar to the traditional FPM, the darkfield measurements lead to spatial resolution improvement in the reconstruction.  To further refine the network, we introduce a mixed loss function that takes a weighted Fourier domain loss, in addition to the standard image domain loss for the generator and the adversarial loss for the discriminator.  We show that this novel weighted Fourier domain loss leads to improved recovery of high frequency information.  We demonstrate our technique using live Hela cell FPM video data from ~\cite{Tian2015b}. We quantitatively assess the performance of our CNN technique over time against those from traditional FPM results, and found that the `generalization' degradation of the reconstructed phase is small over the entire time course (>4 hours).

The training is performed on a PC Intel core i7, 32 GB RAM, NVIDIA GeForce Titan XP for $\sim$16 hours using Keras/Tensorflow framework.  Once the network is trained, reconstructing a 12800$\times$10800 pixels phase image requires only $\sim$25 seconds, which is approximately 50$\times$ faster than the model-based FPM algorithm~\cite{Tian2015b}.

Our technique demonstrates a promising deep learning approach to continuously image large live-cell populations over extended time and gather spatial and temporal information with sub-cellular resolution. Compare to existing FPM~\cite{Zheng2013,Tian2015b}, this CNN approach significantly improves the overall throughput by reducing both the acquisition and computation times, and with less data requirement.  The CNN reconstructed phase image provides high spatial resolution, wide FOV, and low noise-induced artifacts. We also show the flexibility in reconstructing other cell types using transfer learning, which makes our technique appealing to broad applications.

\section{Method}

\subsection{Conditional generative adversarial network (cGAN)}

\begin{figure}[t]
\centering\includegraphics[width=0.8\textwidth]{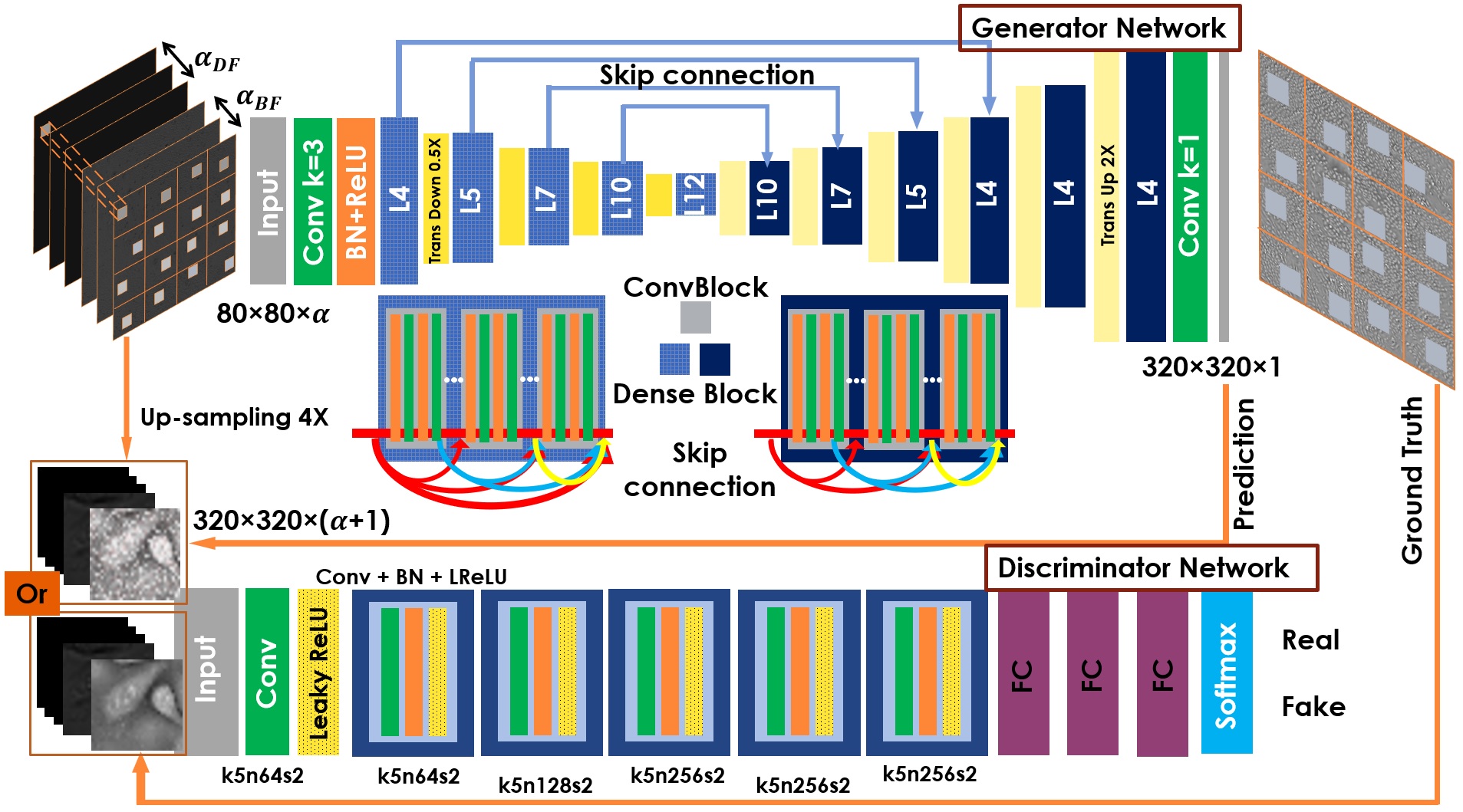}
\caption{The proposed condition generative adversarial network (cGAN) for FPM video reconstruction.  The the generator (top) and the discriminator (bottom) are constructed with the ConvBlock BN-ReLU-Conv($1\times1$)-BN-ReLU-Conv($3\times3$)  and ConvBlock Conv-BN-LeakyReLU, respectively.  
The generator output is the high-resolution phase.  The discriminator tries to  distinguish if that output phase is fake or real.  The generator uses the UNet architecture.  For the discriminator, the generator predicted phase or the ground truth phase is concatenated with the up-sampled intensity data as a conditional input to the discriminator network. The following color schemes are used: the two blocks \protect\img{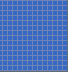} and \protect\img{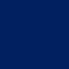} describe the dense concatenation inside the dense block in down-sampling and up-sampling path, respectively. \protect\img{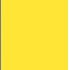} and \protect\img{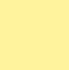} are transition layers interweaving with the dense blocks in the generator. \protect\img{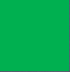} denotes the convolutional layer, \protect\img{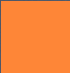} denotes the batch-normalization with a nonlinear ReLU layer in generator model, and \protect\img{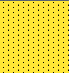} the batch-normalization with the leaky ReLU in the discriminator.  In the last three layers of the discriminator,  \protect\img{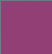} denotes fully-connected layers for high-level feature reasoning. \protect\img{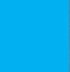} is used at the end for binary classification.   k\#n\#s\# (\# stands for some integer) denotes the filter size, number of channels, and stride of the convolution layer, respectively.}
\label{fig:cGAN}
\end{figure}

Generally speaking, the proposed CNN based FPM reconstruction algorithm takes a set of low-resolution intensity images $I_{\alpha}$ as the network input and output a single high-resolution phase image $\phi_G$.  The intensity images $I_{\alpha}$ are captured from illuminating the sample from $\alpha$ different illumination angles (LEDs) [Fig.~\ref{overview}(a)], in which $\alpha_{\mathrm{BF}}$ are brightfield (BF) and $\alpha_{\mathrm{DF}}$ are darkfield (DF) (Fig.~\ref{fig:cGAN}).  In the training stage, the ground truth phase image is fed into the CNN, which is obtained from the reconstructed high-resolution phase $\phi_{\mathrm{FPM}}$ from the FPM algorithm in~\cite{Tian2015b} [Fig.~\ref{overview}(b)].  A key feature of the FPM is to reconstruct a high-resolution phase image using a set of low-resolution intensity images.  The resolution enhancement factor is $r$ in each dimension.  To obtain the ground truth, it needed to capture the full FPM dataset, containing 173 images~\cite{Tian2015b}.  Since our DL scheme only requires training for the first `FPM frame', the rest of the frame only requires $\alpha (<173)$ images, which allows reducing the acquisition time, especially in a time-series experiment.  We denote the set of $\alpha$  low-resolution images $I_{\alpha}$ as a tensor of dimension $W\times H\times \alpha$ and the corresponding ground truth $\phi_{\mathrm{FPM}}$  a tensor of  dimension $rW\times rH\times1$ [Fig.~\ref{overview}(b)].

The proposed CNN that performs FPM video reconstruction [Fig.~\ref{overview}(c)] is based on the conditional generative adversarial network (cGAN) framework.  It consists of two sub-networks, the generator $G$ and the discriminator $D$ (Fig.~\ref{fig:cGAN}).  Here, the goal of the generator $G$, is to be trained to predict a high-resolution phase $\phi_G = G(I_{\alpha})$ from the low-resolution image set $I_\alpha$ input. To simplify the notation, we will drop the subscript $\alpha$ knowing that $I$ will always contain $\alpha$ low-resolution intensity images. The  generator network $G$ consists of a set of parameters $\theta_G$ (weights and biases), which will be optimized through the training. The optimal $\theta_G$ is learned by minimizing a loss function $l$ over $N$ input-output training pairs:  
\begin{equation}
\widehat{\theta}_{G} = \mathrm{argmin}_{\theta_{G}}\sum_{n=1}^{N}\frac{1}{N}l(G_{\theta_{G}}(I_{n},{\phi_{n}})).
\label{eq:eq1}
\end{equation}
We emphasize that the choice of the loss function $l$ significantly affects the quality of the training.  We propose a mixed loss function that takes the weighted sum of multiple elementary loss functions, which will be detailed in Subsection~\ref{lossfunc}.

The generator $G$ adopts the general "encoder-decoder" architecture used in UNet~\cite{ronneberger2015u} to facilitate efficient learning of pixel-to-pixel information.  UNet has shown to increase the network's performance by adapting to the high-complexity information in image dataset~\cite{simonyan2014very}. To enhance the efficiency of the training process, batch-normalization (BN) is used to offset the internal covariate shift~\cite{ioffe2015batch}. In addition, dropout regularization~\cite{srivastava2014dropout} is employed to constrain network's adaptation to the data during the training to avoid overfitting and increase the network's model accuracy. 
A known problem of training a CNN is that it can get saturated when the network's depth becomes too deep~\cite{he2016deep}. To mitigate this problem, the dense block (DB) proposed in the densely connected network is used~\cite{huang2017densely}.  
 A DB connects each layer to its subsequent layers in a feed-forward fashion.  The inputs to each layer are the feature-maps of all preceding layers; the output of the current layer's own feature-maps are inputs to all the subsequent layers (see Fig.~\ref{fig:cGAN}). 
The DB has several advantages, including (a) mitigation of the vanishing-gradient problem in the training; (b) reduction of the total number of parameters; (c) enhancement of feature propagation and reuse.  
 A typical $L$-layer DB is defined as follows:
\begin{equation}
x_{L} = H_{L}([x_{0},x_{1},\cdots,x_{L-2},x_{L-1}]).
\label{eq:eq2}
\end{equation}
where $[\cdot]$ denotes the concatenation operation that connects all the feature maps of all $L$ layers in the block.  The output at the end of each $L$-layer DB $H_L(\cdot)$  has  $R_0 + R\times(L-1)$ numbers of feature maps, where $R_0$ is the number of the feature maps in the first layer, the hyper-parameter $R$ is referred to as the growth rate.   Within each layer  inside the DB (ConvBlock), a series of operations are performed, including batch-normalization (BN), nonlinear activation using the ReLU or LeakyReLU function (ReLU/Leaky ReLU)~\cite{agostinelli2014learning}, and convolution with filters of kernel size $k\times k$ [Conv($k\times k$)].

In our generator $G$,  it contains a total of 11 DBs.  The number of ConvBlock layers in each DB is $L=[4, 5, 7, 10, 12, 10, 7, 5, 4, 4, 4]$ (marked as $L\#$ in Fig.~\ref{fig:cGAN} with $\#$ denoting the number of ConvBlock layers in each DB).  In each ConvBlock layer, a stack of BN-ReLU-Conv($1\times1$)-BN-ReLU-Conv($3\times3$) operations are performed with $R=12$ and $R_0 = 46$.

Between two consecutive DBs, a transition block is used to facilitate the desired down-sampling or up-sampling operation.  The down-sampling transition block contains Conv($1\times1$)-BN-ReLU-Conv($3\times3$, stride=2); the up-sampling transition block contains Conv($1\times1$)-BN-ReLU-Deconv($3\times3$, stride=2), where Deconv denotes the deconvolution (transpose convolution) layer~\cite{dumoulin2016guide}.
The features of the input layer are extracted by an initial Conv($3\times3$)-BN-ReLU block before feeding them to the first DB. A Conv($1\times1$) is used to perform the final regression to generate the phase map $\phi_G$.   

The discriminator network $D$ aims to distinguish if the output from $G$ is real or fake.  
Following \cite{goodfellow2014generative} and \cite{isola2017image}, we define a conditional Generative Adversarial Network (cGAN) to solve the following adversarial min-max problem: 
\begin{align}
min_{\theta_{G}}max_{\theta_{D}}E_{I,\phi}[logD_{\theta_{D}}(I,\phi)]+ 
E_{I}[log(1-D_{\theta_D}(I,G(I))].
\label{eq:eq3}
\end{align}

The general idea behind this network is that it aims to train a generator $G$ to `fool' the discriminator $D$.  Here, $D$ is trained to distinguish whether the high-resolution phase image predicted by $G$ represents a real phase image.  It was observed that GAN in general is hard to train and it may fail when the generator collapses to a parameter setting where it always gives the same output. A successful strategy to avoid this failure is to allow the discriminator to perform minibatch discrimination  \cite{isola2017image,salimans2016improved}. In this case, the discriminator distinguishes if the reconstructed phase image is real or fake by evaluating multiple sub-regions of the $G$-predicted image instead of the whole.

\subsection{Loss function}
\label{lossfunc}

A motivation of the usage of the discriminator network $D$ is that the commonly used pixel-wise loss functions, such as the mean absolute error (MAE), mean square error (MSE), and structural similarity index (SSIM), may not be the most appropriate figures of merit, in particular when assessing a CNN's performance in preserving high frequency content of  reconstructed images.  
The minimization of these pixel-wise loss functions can lead to solutions that ignores the high-frequency details, while favors solutions that are smooth, albeit have less perceptual quality~\cite{wang2004image}. With cGAN approach, the generator $G$ can learn to create a solution that resembles realistic high-resolution images with high-frequency details. 

For this purpose, we define the `perceptual loss function' $l$ as a weighted sum of multiple loss functions. This ensures that the model can learn the desired features containing both low-frequency and high-frequency information in the phase images. 
Specifically, our loss function consists of four components, including the pixel-wise  spatial domain mean-absolute error (MAE) loss  $l_{\mathrm{MAE}}$, the pixel-wise Fourier domain mean-absolute error (FMAE) loss $l_{\mathrm{FMAE}}$, the generator's adversarial loss $l_G$, and the weight regularization $l_{\theta_G}$,  in the following form:

\begin{equation}
\label{eq:eq4}
    l= \lambda_1(\beta_1l_{\mathrm{MAE}}+\beta_2l_{\mathrm{FMAE}})+\lambda_2l_G+\lambda_3l_{\theta_G},
\end{equation}
where
\begin{align}
\begin{split}\label{eq:eq5}
    l_{\mathrm{MAE}} ={}&\frac{1}{r^2WH}\left\||\phi|-|G_{\theta_G}(I)|\right\|,
\end{split}\\
\begin{split}\label{eq:eq6}
l_{\mathrm{FMAE}} ={}&\frac{1}{r^2WH}\left\||\ft(\phi)|-|\ft(G_{\theta_G}(I))|\right\|,
\end{split}\\
\begin{split}\label{eq:eq7}
l_G ={}&-\log D_{\theta_G}(I,G(I)),
\end{split}\\
    l_{\theta_G} ={}& \|\theta_{G}\|,
    \label{eq:eq8}
\end{align}
where $\ft$ denotes the 2D Fourier transform, $\|\cdot\|$ is the $L_1$-norm.
$(\lambda_1,\beta_1,\beta_2,\lambda_2,\lambda_3)$ are hyper-parameters that controls the relative weights of each loss components. 
We found that the Fourier loss function is sensitive to pixel-wise corruption during the early stage of the training process.  As a result, we use it only to refine the outputs by enforcing similarity in the frequency domain~\cite{yang2017dagan} after initial training is done with the other three loss components (details in Subsection~\ref{training}).

\subsection{Data preparation}

To test our CNN technique, we use FPM video data from~\cite{Tian2015b}.  The time-series data was taken on Hela cells at 2 min intervals over the course of $\sim$4 hours that contains several cell cycles. Each FPM dataset contains 173 low-resolution intensity images, in which 37 are brightfield (BF), 136 are darkfield (DF).  Each intensity image is 2560$\times$2160 pixels in 16-bit grayscale. 

To generate the data for training, FPM phase reconstructions from \cite{Tian2015b} are used as the ground truth.  Each FPM reconstructed phase image contains 12800$\times$10800 pixels, which is 5$\times$5 larger than the raw intensity image. 

To prepare the dataset for training, we  use only the first FPM frame in the time-lapse as the training set.  Specifically, to prepare the ground truth data, the full FOV phase image is first divided into 4$\times$4 sub-regions, containing 3440$\times$2760 pixels.  To avoid edge artifacts during training and reconstruction, neighboring sub-regions are chosen to have 320-pixel and 80-pixel overlap along the horizontal and vertical directions, respectively.  The corresponding intensity image in each sub-region are with 688$\times$552 pixels.  The input to the CNN are BF and DF image patches that are cropped from random locations  of each of the sub-region images, each with 64$\times$64 pixels.  Each training input data is formed by stacking the BF and DF image patches to form a 64$\times$64$\times\alpha$ tensor.  To facilitate fast computation, the models are designed with down-sampling path and up-sampling path. Each input image was up-sampled
to 80 $\times$ 80 using bilinear interpolation. The spatial dimension of each layer in the CNN are 80, 40, 20, 10, 20, 40, 80, 160 and 320, respectively.  The corresponding ground truth data contains 320$\times$320 pixels.  Each raw BF image was preprocessed by the background subtraction procedure in \cite{Tian2015b}; each raw DF image was preprocessed to remove the dark current noise~\cite{Tian2015b}. The same preprocessing steps are applied for training, validation, and testing.

\subsection{Training, evaluation, and testing}
\label{training}

\begin{figure*}[t]
\centering\includegraphics[width=0.75\linewidth]{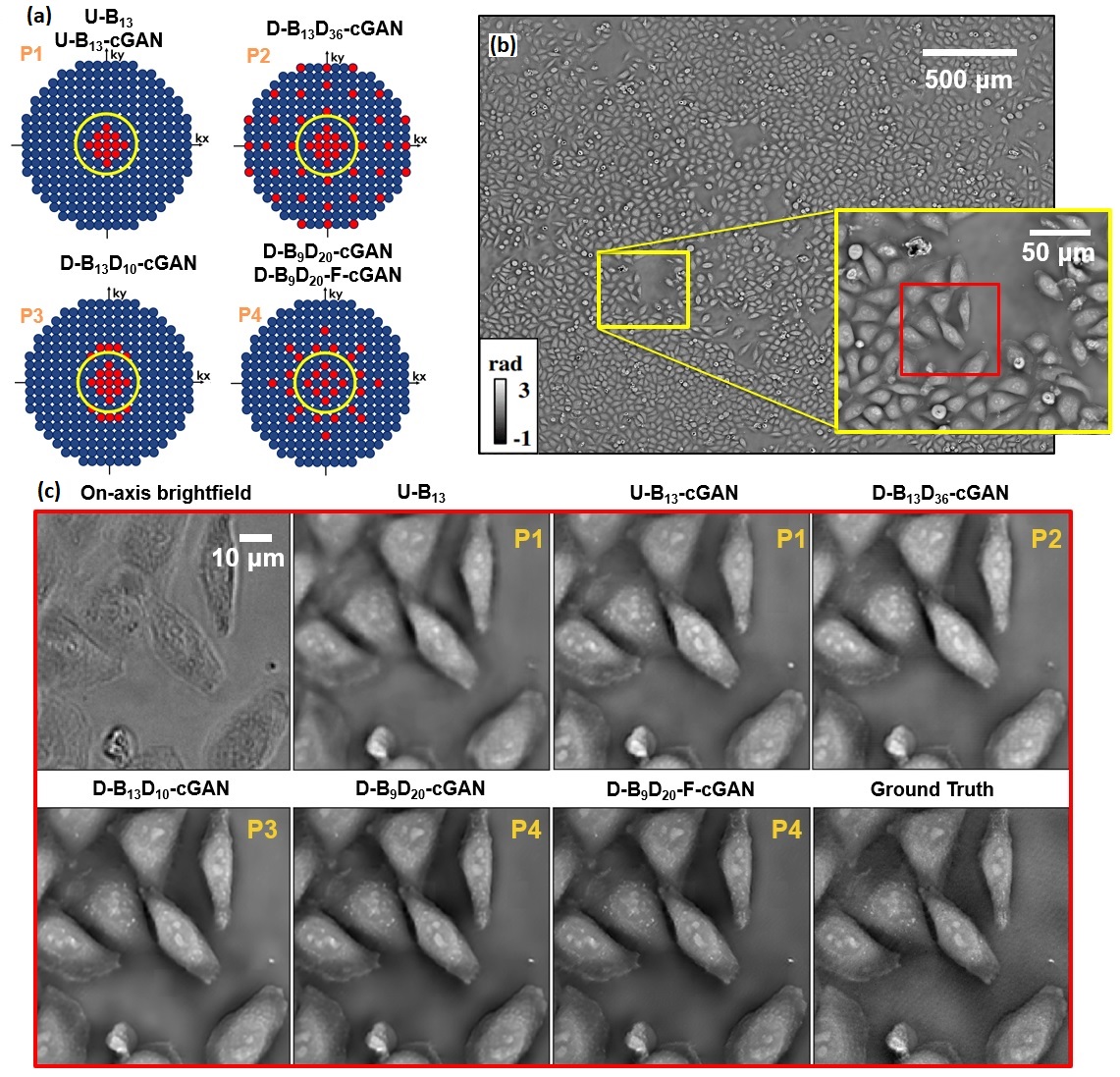}
\caption{(A) The summary of the illumination patterns and network structures investigated. The illumination angles (shown in the Fourier space) in-use are marked in red.  The yellow cycle indicates the NA of the imaging system. (B) A sample full-FOV high-SBP phase reconstruction (at 4 hour) predicted by the proposed network D-B$_{9}$D$_{20}$-F-cGAN. (C) The original intensity image, ground truth phase image, and the reconstructions from the CNN models from the zoom-in area [marked by the red square in (B)].}
\label{fig:patternAndROI}
\end{figure*}

To investigate the interplay between the illumination pattern and the performance of the CNN, we train our network by using several different combinations of BF and DF images.  The illumination patterns along with the CNN models used are shown in Fig.~\ref{fig:patternAndROI}(a).  Each illumination pattern is plotted in the Fourier space, in which a yellow circle indicates the NA of the objective lens.  Intensity images taken from the LEDs within the circle are BF; whereas those outside the circle are DF.  The LEDs in-use are marked in red. To systematically study the relation between the reconstructed resolution with the illumination's angular coverage, we have designed patterns with (P1) 13 BF-only with 0.2 illumination NA, (P2) 13 BF + 36 DF with 0.6 illumination NA, (P3) 13 BF + 10 DF with 0.25 illumination NA, and (P4) 9 BF + 20 DF with 0.4 illumination NA.  The following networks are investigated: 
P1 is trained on two networks,  U-B$_{13}$ implements the UNet without DB in~\cite{nguyen2017automatic}; U-B$_{13}$-cGAN implements the UNet in~\cite{isola2017image} with the cGAN architecture (i.e. with the discriminator network $D$ in Fig.~\ref{fig:cGAN}); 
P2 is trained on the cGAN network in Fig.~\ref{fig:cGAN}, D-B$_{13}$D$_{36}$-cGAN;  P3 is trained on the cGAN network, D-B$_{13}$D$_{10}$-cGAN; P4 is trained on a  cGAN network with and without the Fourier loss function, denoted as D-B$_{9}$D$_{20}$-cGAN, D-B$_{9}$D$_{20}$-F-cGAN, respectively. 

Each model was trained with $\sim$700-900 epochs.  For UNet, the batch-size was 16; whereas the batch-size was 4 in UNet with DB due to memory limitation. We use the weight coefficients $(\lambda_1,\beta_1,\beta_2,\lambda_2,\lambda_3)=(10^2, 1, 0, 1, 10^{-5})$ when the Fourier loss is not used.  When the Fourier loss is used, we first train the network with $(\lambda_1,\beta_1,\beta_2,\lambda_2,\lambda_3)=(10^2, 1, 0, 1, 10^{-5})$ for 700 epochs, and then with $(\lambda_1,\beta_1,\beta_2,\lambda_2,\lambda_3)=(10^2, 0.95, 0.05, 1, 10^{-5})$ for another $145$ epochs.  We observed that the network's parameters are unstable in the early stage of training.  To stabilize the training process, we  added the Fourier loss after 700 epochs. We used the ADAM optimizer~\cite{kingma2014adam} with initial learning rate of $10^{-5}$, dropout factor 0.5 after every 10 epochs, in which each epoch contains 1000 iterations. In each iteration, the algorithm incrementally updates the model using a subset (set by the batch-size) of the input. To fine tune each network, as an optional step, we performed model validation using the FPM frame taken at 2 hour.  The best models were selected based on the MAE metric calculated on the validation data.  

Once the CNN is trained, which only needs to be performed once using the first FPM frame taken at 0 min, the CNN is then applied to reconstruct high-SBP phase video frames (i.e. the testing step).  To perform the reconstruction, similar data preprocessing steps are followed as the training phase.  The raw intensity images were first divided into 4 $\times$ 4 sub-regions.  Within each region, image patches having the same sizes as training batches (64$\times$64$\times\alpha$) are used for reconstruction.  Neighboring image patches contain 15-pixel and 19-pixel overlap in the horizontal and vertical directions, respectively. Each image patch was first up-sampled to 80$\times$80 pixels with bilinear interpolation. The predicted phase image contains 320 $\times$ 320 pixels. Once reconstructions are performed on all 2288 patches, the alpha blending algorithm was used to form the full FOV phase image containing 12800 $\times$ 10800 pixels.  To reconstruct the video, we simply fed each FPM frame to the trained CNN to reconstruct the high-SBP dynamic information from the times-series data.  The time for reconstructing each full-FOV, high-SBP image is $\sim$25$\pm$2 seconds using our cGAN network with the added Fourier loss, D-B$_{9}$D$_{20}$-F-cGAN, which is $\sim$50$\times$ faster than the standard FPM algorithm (which took $\sim$20 min for each frame~\cite{Tian2015b}). A detailed comparison of all networks is detailed in Section~\ref{sec:results} and Table~\ref{tab:metrics1}.

\begin{figure*}[t]
\centering\includegraphics[width=0.685\textwidth]{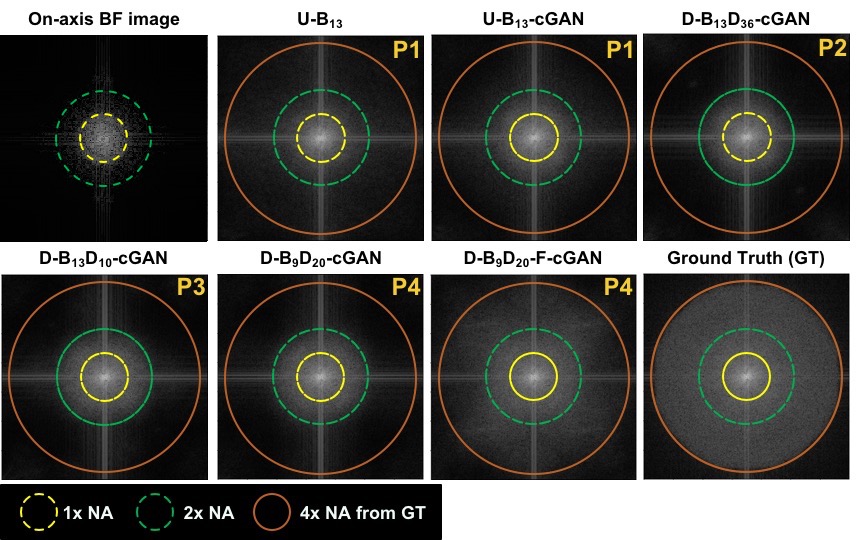}
\caption{Fourier analysis of the CNN reconstructed phase images. We directly take the Fourier transform of the reconstructions in Fig.~\ref{fig:patternAndROI}(c). They are compared with the raw intensity image from on-axis illumination and the ground truth from FPM. To illustrate the Fourier coverage in each model, we mark three circles in each image, in which the yellow circle corresponds to the support of the pupil function with a radius of $1\times$NA, the green circle corresponds to the support of the optical transfer function with a radius of $2\times$NA, and the orange circle is the support from the ground truth with a radius of $4\times$NA.}
\label{fig:FM}
\end{figure*}

\begin{figure*}[ht]
\centering\includegraphics[width=0.75\linewidth]{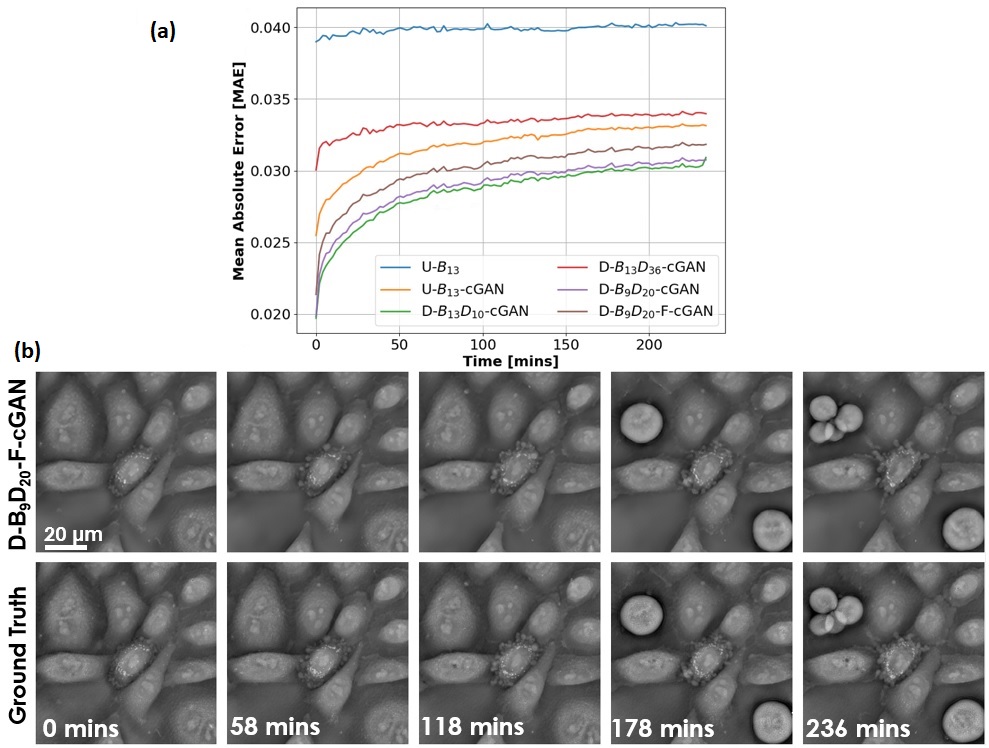}
\caption{Reconstructed temporal dynamic information using the proposed CNN. (A) The MAE metric is evaluated for every frame of the time-series experiment on all the CNN models. (B) Several frames of the reconstructed high-SBP phase video (see \href{https://figshare.com/s/b6a7fc88b22edfd1af28}{Visualization 1} for more examples) from a zoom-in region, where significant morphological changes are observed over the course of 4 hours.}
\label{fig:FramMAEandTimelapse}
\end{figure*}

\section{Results and discussion}
\label{sec:results}

We discuss our results by presenting results in real space (Fig.~\ref{fig:patternAndROI}), Fourier space (Fig.~\ref{fig:FM}), and over different time points (Fig.~\ref{fig:FramMAEandTimelapse}).

Figure~\ref{fig:patternAndROI}(a) summarizes all the illumination patterns used for training and testing along with the corresponding networks used. All networks are applied to reconstruct the entire time-series experiment.  A sample large-SBP phase reconstruction across the full (1.7mm$\times$2.1mm) FOV is shown in Fig.~\ref{fig:patternAndROI}(b).  In Fig.~\ref{fig:patternAndROI}(c), we zoom-in on a sub-region to compare the results from different networks in the real space.  For comparison, the raw low-resolution intensity image from the  central BF illumination is shown, which was bilinearly up-sampled to have the same size as the network's output.  

The result from U-B$_{13}$, which uses BF data only, UNet without DB or cGAN and only the pixel-wise MAE loss, produce low-resolution phase images.  It has been shown that the MAE loss function can lead to blurry results when solving an image reconstruction problem~\cite{isola2017image} because it does not place sufficient weights in the high frequency content.  
To overcome this problem, we use generative adversarial networks to reconstruct phase image with conditional input (cGAN)~\cite{goodfellow2014generative}.  In U-B$_{13}$-cGAN, the UNet is accompanied with a discriminator network in order to better learn high frequency information.  
The introduction of the cGAN architecture allows us to better reconstruct sub-cellular structures with more perceptual details; however, the resolution still appears worse compared to the ground truth.  

In order to further improve resolution, as in FPM, DF images are needed since they contain high-spatial frequency information beyond the support of the optical transfer function (OTF).  In addition, to deal with the added data size, we also seek a more efficient network structure with higher representation power.  The dense block (DB) structure has shown to provide efficient presentation with a small number of parameters in the model~\cite{huang2017densely}.  We present results from three illumination patterns with different angular coverage, all reconstructed with DenseNet (UNet with DB) and cGAN structure.   In D-B$_{13}$D$_{36}$-cGAN, we use 36 DF images covering up to 0.6 illumination NA.  
This leads to moderate resolution improvement; however, the results are limited by the highly noisy data captured at very large NAs.  
In general, we observe that it is not guaranteed that higher illumination angle leads to better resolution.  The reason is because the DF data is subject to much higher level noise than the BF data, and the noise level increases as the the illumination angle increases~\cite{Yeh2015}.  When the signal-to-noise ratio (SNR) falls below certain threshold, the inclusion of these DF data is no longer helpful.  
To confirm this, we first use a small amount of DF data from small angles in D-B$_{13}$D$_{10}$-cGAN.  It leads to resolution improvement as compared to the one from D-B$_{13}$D$_{36}$-cGAN.  
It should be noted that the DF SNR can significantly improve if a dome-shaped LED array~\cite{Phillips2015} is used instead of the planar array in \cite{Tian2015b}. 
Heuristically, we found that the capacity of our CNN is that it can reliably utilize DF data up to 0.4 illumination NA (P4).   The reconstructions are further explored using two networks, D-B$_{9}$D$_{20}$-cGAN and D-B$_{9}$D$_{20}$-F-cGAN.  

A major limitation of image-space only loss function is that the metric still favors low-frequency information~\cite{wang2004image} but under-weights high-frequency information.  A recently proposed solution is to further include Fourier loss component~\cite{yang2017dagan}.  The result from using this strategy is shown in D-B$_{9}$D$_{20}$-F-cGAN. Our reconstruction of last frame on Hela cells is available at \cite{helaDLFPMGit}. 

\begin{table}[t]
\caption{\bf Performance metrics evaluated on the full-FOV testing data [Legend: * stands for -cGAN, $\bullet$ based on the region in Fig. \ref{fig:patternAndROI}(c), GT: ground truth]}
\centering
\begin{tabular}{ccccccc}
\hline
Method		          &MAE    &PSNR     &SSIM    &FM$\bullet$ & Time     \\ \hline
U-B$_{13}$        &0.0401 & 25.01db & 0.7575 & 0.0110   & 30$\pm$3s \\ \hline
U-B$_{13}$* &0.0331 & 26.49db & 0.7790 & 0.0156   & 55$\pm$3s \\ \hline
D-B$_{13}$D$_{36}$* &0.0339 & 26.17db & 0.7779 & 0.0146   & 25$\pm$2s \\ \hline
D-B$_{13}$D$_{10}$* &0.0309 & 26.76db & 0.7966 & 0.0165   & 25$\pm$2s \\ \hline
D-B$_{9}$D$_{20}$*  &0.0308 & 26.87db & 0.7964 & 0.0169   & 25$\pm$2s \\ \hline
D-B$_{9}$D$_{20}$-F* &0.0318 & 26.19db & 0.7797 & 0.0211   & 25$\pm$2s \\ \hline
FPM (GT)               &0      & 1.00    & 0      & 0.0389   & 20 mins~\cite{Tian2015b}  \\ \hline
\end{tabular}
  \label{tab:metrics1}
\end{table}

To better visualize the recovery of high-frequency information, Fig.~\ref{fig:FM} shows the Fourier transform of each image in Fig.~\ref{fig:patternAndROI}(c).  The spectrum of the on-axis BF image is mostly concentrated within the pupil region, i.e. the circular region with a radius of 1$\times$NA, and extends up to the support of the OTF (i.e. 2$\times$NA).  It is well known that using only BF images can provide Fourier coverage up to the support of the OTF.  As shown in the Fourier image of U-B$_{13}$-cGAN, the network is able to fully recover these low-frequency information. The inclusion of DF images {\it should} lead to larger Fourier coverage; however, the improvement is not significant with image-space only loss function, as shown in the Fourier images of D-B$_{13}$D$_{36}$-cGAN, D-B$_{13}$D$_{10}$-cGAN, and B$_{9}$D$_{20}$-cGAN.  The introduction of the Fourier domain loss significantly boosts the Fourier coverage up to the 0.4 illumination NA (<0.6 NA in the ground truth), as shown in the Fourier image of D-B$_{9}$D$_{20}$-F-cGAN.  We note that using the Fourier domain loss in the training process generally leads to enhancement of the sharpness of the results and the frequency measurement metric (FM)~\cite{de2013image}; however, it may trade off image-space metrics, such as MAE, SSIM, and PSNR due to different metric weighting schemes involved (see Table.~\ref{tab:metrics1}).  

Further inspecting the results from the CNN and comparing them to the FPM generated `ground truth', we note that the ground truth image contains noisy structures, which are clearly visible in the background.  All CNN reconstructed results are free from these background artifacts, demonstrating the robustness of the training process to noisy ground truth data. 

A unique feature of our technique is the ability to reconstruct high-SBP phase videos with  training data only from the first time point of a long time-series experiment.  
To demonstrate the effectiveness of this strategy, we show our CNN predicted temporal frames over the course of over 4 hours. 
During this process, considerable amount of morphological (hence phase distribution) changes occur due to cell division over several cell cycles.  Figure~\ref{fig:FramMAEandTimelapse}(b) shows several frames (reconstructed with D-B$_{9}$D$_{20}$-F-cGAN) of a zoom-in region, where one cell is growing and dividing into multiple cells, and another cell has its membrane rapidly fluctuating.  More example videos are provided in \href{https://figshare.com/s/b6a7fc88b22edfd1af28}{Visualization 1}.
A more quantitative evaluation of the `generalization error' over time is presented in Fig.~\ref{fig:FramMAEandTimelapse}(a), in which the MAE metrics of all the networks studied are plotted for every frame in the time series experiment.  The error is low at the beginning of the experiment and grows slowly as the time progresses.

\begin{figure*}[ht!]
\centering\includegraphics[width=0.95\textwidth]{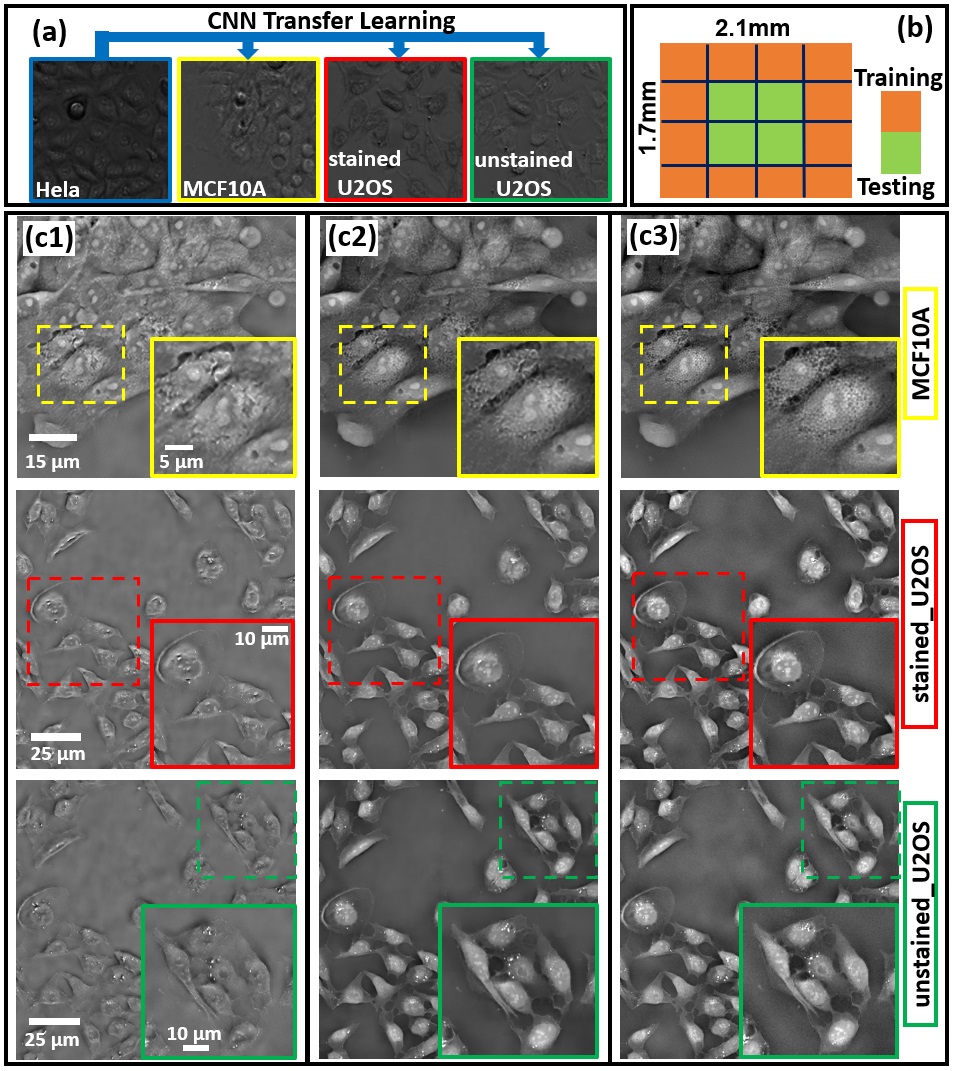}
\caption{Transfer learning using the pre-trained CNN (D-B$_{9}$D$_{20}$-F-cGAN) on Hela cells, and then used to make predictions of the phase on MCF10A, stained and unstained U2OS cells. (a) the intensity images vary across different cell types and before/after staining. The image patches are taken from the same FOV region and using the same illumination angle. (b) The regions used for testing and training for demonstrating the transfer learning.
Phase reconstructed from (c1) directly apply the pre-trained CNN to the new data. (c2) after 30min transfer learning. (c3) the ground truth from~\cite{Tian2015b}.}
\label{fig:TL}
\end{figure*}

\section{Transfer learning}
Practically, it is difficult to train a single network that can handle all sample types, a main drawback of the DL approach compared to the model based methods. 
To mitigate this problem, we investigate transfer learning, in which our pre-trained CNN on Hela cell is finely tuned for other cell types. The effectiveness of this strategy to address the generalization limitation of sample types has also been previously demonstrated in other biomedical imaging applications~ \cite{ozcan2018Deep_histology}.

We used D-B$_{9}$D$_{20}$-F-cGAN trained on Hela cells to predict the phase reconstruction of two other cell types (MCF10A, U2OS) with or without staining. 
The data were captured with the same setup in~\cite{Tian2015b}. 
In Fig.~\ref{fig:TL}, we compare two results.  
First, we directly apply the D-B$_{9}$D$_{20}$-F-cGAN network to the new data.
To further refine the results, we use the transfer learning technique.  
Specifically, we take the weights from the pretrained network and continue the training with the new cell data as the training data for $\sim$30 mins.  
Note that these new cell data contain significant intensity differences. 
By fine tuning the model, the CNN is able to produce high quality reconstruction. 
During the transfer learning, we did not use any validation data and only evaluated the new CNN's performance directly after the 30-min training. 
The results show that transfer learning provides a practical way to broaden the utility of our technique. 

\section{Conclusion}
We have demonstrated a deep learning framework for Fourier ptychography video reconstruction.  The proposed CNN architecture fully exploits the unique high-SBP imaging capability of FPM so that it can be trained using a single frame and then be generalized to a full time-series experiment.  In addition, the CNN requires reduced number of images for high-resolution phase recovery.  The reconstruction of each high-SBP image takes less than 30 seconds.  Overall, this technique significantly improves the imaging throughput of the FPM system by reducing both the acquisition and reconstruction time.  The central idea of our technique is based on the observation that each FPM image contains a large cell ensembles covering all morphological information throughout the time-series experiment.  By the principle of ergodicity, the statistical information learned from these large spatial ensembles in a single frame are shown to be sufficient to predict temporal dynamics with high fidelity. In practice, we showed that our trained CNN can successfully reconstruct a high-SBP phase video of dynamic live cell populations with reduced noise artifacts. Using the conditional generative adversarial network (cGAN) framework and a weighted Fourier loss function, the proposed CNN is able to more effectively learn the high resolution information encoded in the darkfield data.  The technique may find wide applications in {\it in vitro} live cell imaging  and gather large-scale spatial and temporal information in a data and computation efficient manner. We also demonstrate that transfer learning is a practical approach to image a broad range of new cell samples, bypassing the need to train an entirely new CNN from scratch.

\section*{Acknowledgments}
We would like to thank NVIDIA Corporation for supporting us with the GeForce Titan Xp through the GPU Grant Program.

\section*{Disclosures}
The authors declare that there are no conflicts of interest related to this article.

\end{document}